\documentclass[letterpaper, 10 pt, conference]{ieeeconf} 
\IEEEoverridecommandlockouts
\usepackage{cite}
\usepackage{amsmath,amssymb,amsfonts}
\usepackage{algorithmic}
\usepackage{graphicx}
\usepackage{textcomp}
\usepackage{xcolor}
\usepackage{hyperref}
\usepackage[caption=false,font=normalsize,labelfont=sf,textfont=sf]{subfig}

\title{\LARGE \bf Introducing V-Soft Pro: a Modular Platform for a Transhumeral Prosthesis with Controllable Stiffness}

\author{Giuseppe Milazzo, Giorgio Grioli, Antonio Bicchi, and Manuel G. Catalano%
\thanks{This work was supported by the European Union’s ERC Synergy Natural BionicS under Grant ID 810346. The content of this publication is the sole responsibility of the authors. The European Commission or its services cannot be held responsible for any use that may be made of the information it contains.}
\thanks{Corresponding author: Giuseppe Milazzo. email: giuseppe.milazzo@iit.it}
\thanks{All the authors are with Soft Robotics for Human Cooperation and Rehabilitation, Istituto Italiano di Tecnologia, Genova 16163, Italy.}
\thanks{G. Grioli and A. Bicchi are also with the Research Center “Enrico Piaggio” \& Department of Information Engineering, University of Pisa, Pisa 56126, Italy.}
\thanks{This article has supplementary downloadable material available at \url{https://ieeexplore.ieee.org}, provided by the authors.}%
}

\begin{document}
\maketitle
\thispagestyle{empty}
\pagestyle{empty}

\begin{abstract}
Current upper limb prostheses aim to enhance user independence in daily activities by incorporating basic motor functions. However, they fall short of replicating the natural movement and interaction capabilities of the human arm. In contrast, human limbs leverage intrinsic compliance and actively modulate joint stiffness, enabling adaptive responses to varying tasks, impact absorption, and efficient energy transfer during dynamic actions. Inspired by this adaptability, we developed a transhumeral prosthesis with Variable Stiffness Actuators (VSAs) to replicate the controllable compliance found in biological joints.
The proposed prosthesis features a modular design, allowing customization for different residual limb shapes and accommodating a range of independent control signals derived from users' biological cues. Integrated elastic elements passively support more natural movements, facilitate safe interactions with the environment, and adapt to diverse task requirements. This paper presents a comprehensive overview of the platform and its functionalities, highlighting its potential applications in the field of prosthetics.
\end{abstract}

\section{Introduction}\label{sec:Introduction}
Limb loss brings profound challenges to individuals, affecting not only their motor capabilities but also their emotional well-being and social lives. Prosthetic devices help bridge this gap by restoring the natural appearance of the human body and some functional capabilities of human limbs. However, current prostheses fall short of replicating the complex functionality of natural limbs, focusing mainly on enabling users to achieve a basic level of self-sufficiency with essential motor functions.

Myoelectric upper-limb prostheses represent the current frontier in human-like limb restoration and natural motor control.
However, robotic prostheses often rely on rigid joints, which lack the fluid, adaptive behavior that human limbs exhibit in motion and interaction with the environment.
Humans voluntarily adjust the impedance of their limbs by coactivating antagonistic muscles.
This adaptive strategy provides flexible and responsive limb behavior: a compliant approach reduces applied force during constrained tasks and allows safe exploration in unfamiliar environments, while high stiffness enables precision tasks and counters external forces effectively~\cite{osu2004optimal,borzelli2018muscle,phan2020estimating,blank2014task}. Inspired by this capability, it is clear that upper-limb prosthetic devices should incorporate mechanisms for users to actively control impedance, which would enhance both versatility and task performance~\cite{blank2014task, hogan1983prostheses}.
Unlike human limbs, rigid joints lack the adaptability provided by passive compliance and controllable stiffness. Rigid devices are also less suitable for safe interaction in unstructured and dynamic environments, where humans frequently face unpredictable conditions in daily life. Unable to absorb shocks, these devices increase the risk of harm to people and objects during unintended collisions, which are common due to a lack of proprioceptive feedback. Furthermore, rigid prostheses are limited in their ability to apply low interaction forces, making delicate interactions with fragile objects challenging.
\begin{figure}[!t]
    \centering
    \subfloat[]{\includegraphics[width=0.49\linewidth]{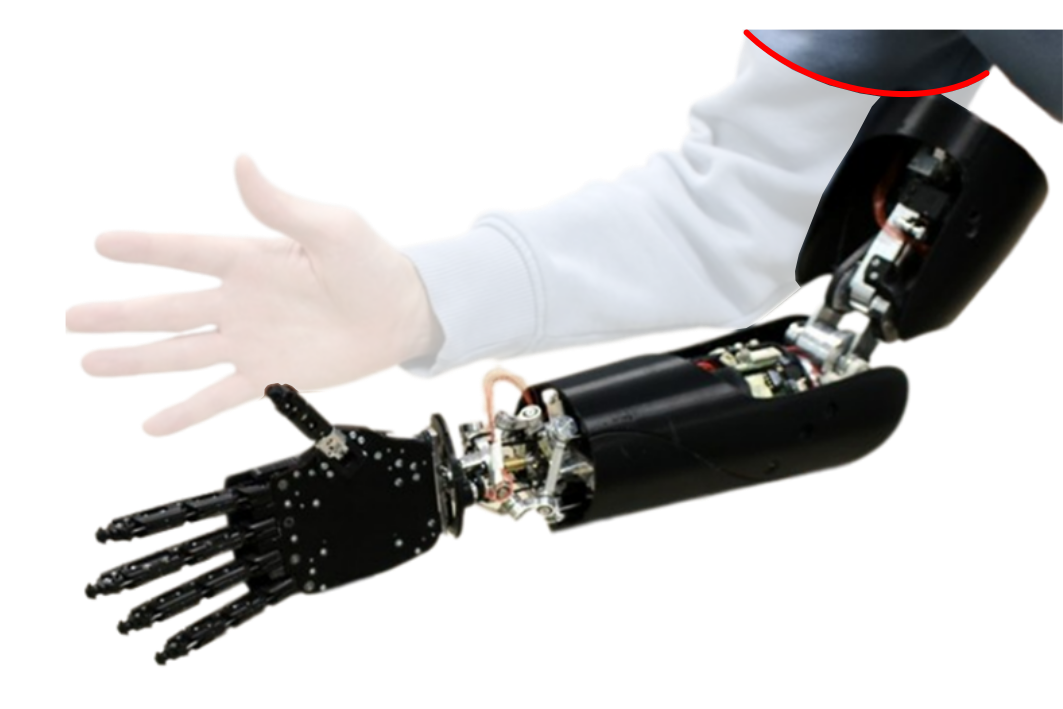}\label{fig:Fig1a}}
    \hfill
    \subfloat[]{\includegraphics[width=0.49\linewidth]{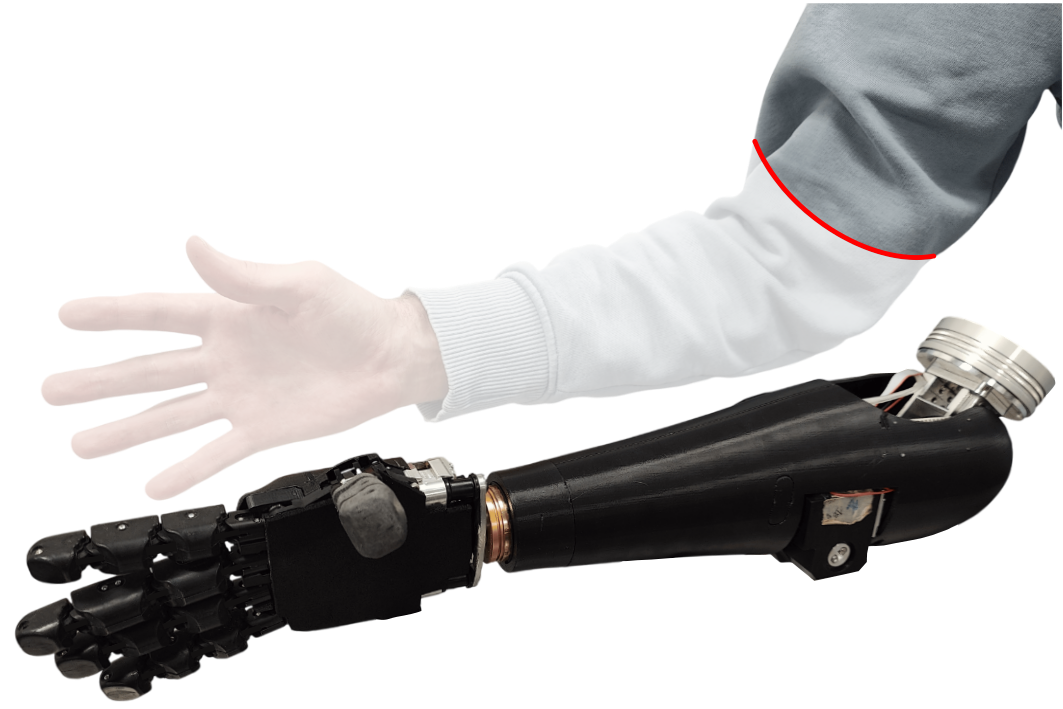}\label{fig:Fig1b}}
    \caption{Two configurations of the V-Soft Pro transhumeral prosthesis, customized to meet diverse user needs. Panel (a) illustrates a V-Soft Pro configuration designed to achieve a human-like weight distribution for proximal transhumeral amputations. Panel (b) shows a version of the V-Soft Pro platform adapted for users with longer residual limbs.}\label{fig:Fig1}
\end{figure}

Impedance control has revolutionized robotic applications that require physical interaction with the environment and cooperation with humans~\cite{hogan1985impedance, haddadin2024unified,zhao2023human,liao2024trust}, enabling robots to follow planned paths during free motion and supervise interaction forces when encountering obstacles along the path.
Impedance control can be implemented in standard torque-controlled robots simply via software. However, it relies on high-performance sensors, actuators, and computing, making it well-suited for most industrial applications but difficult to integrate into portable robotic devices with limited battery life and performance, such as prostheses~\cite{english1999mechanics}.
An alternative approach is to embed variable impedance characteristics directly into the robot hardware design~\cite{howard2013transferring, tagliamonte2012double, tiboni2022sensors}. On one hand, Variable Impedance Actuators (VIAs) involve complex hardware implementations, as they require dedicated mechanisms to alter the physical properties of the joint. On the other hand, they often rely on simple software control and are intrinsically robust and versatile, as joint impedance is dictated passively by the physical properties of the joint. Among VIAs, VSAs integrate redundant actuation and nonlinear elastic transmissions into their mechanical design, allowing them to adjust output stiffness by controlling the active deflection of elastic elements, akin to antagonistic muscle coactivation.

This paper introduces V-Soft Pro (V-SP), a modular platform for an active transhumeral prosthesis with user-controllable stiffness. The platform consists of two elbow modules, three wrist modules, and two hand modules, which can be combined to create a transhumeral prosthesis tailored to the user's limb morphology and the available biosignals for control. While some of the V-SP modules have been previously presented and characterized in our past works~\cite{milazzo2024elbow,milazzo2024modeling}, this paper focuses on their integration into a customizable transhumeral prosthesis with variable stiffness (VS). Section~\ref{sec:Related_Work} provides an overview of the state of the art in transhumeral prostheses and VS prosthetic joints. Section~\ref{sec:V-SoftPro} details the V-SP hardware design and the implementation of its submodules. Section~\ref{sec:Experiments} presents the hardware integration of two different configurations of the V-SP platform and demonstrates the functionality of the VS transhumeral prostheses. Section~\ref{sec:Discussion} discusses potential applications and limitations of the system, and conclusions are provided in Section~\ref{sec:Conclusion}.

\section{Related Work}\label{sec:Related_Work}
Robotic upper limb prostheses typically feature rigid actuation and fixed impedance properties. However, some elbow and wrist prostheses offer a transition between two discrete stiffness levels (locked or unlocked) by activating a locking mechanism through electromyographic (EMG) control or with the contralateral arm. For instance, elbow prostheses\footnote{\label{note1}\href{https://www.utaharm.com/product-utah-arm-options/}{Fillauer Utah Arm U3}}${}^\text{,}$\footnote{\label{note2}\href{https://shop.ottobock.us/Prosthetics/Upper-Limb-Prosthetics/Myoelectric-Elbows/DynamicArm/p/12K100N~550-1}{Ottobock Dynamic Arm}} can be locked to support external weights or unlocked to swing freely during walking~\cite{bennett2016design}. Similarly, spring-loaded flexion wrists\footnote{\label{note3}\href{https://irp-cdn.multiscreensite.com/acf635e2/files/uploaded/Flexion\%20Wrist.pdf}{Touch Bionics Flexion Wrist}} can be unlocked to exploit environmental constraints during grasping or locked to resist perturbations~\cite{capsi2024three}.
While these locking mechanisms provide some degree of task adaptability, they lack the natural ability to smoothly and accurately control joint stiffness.
VSAs have recently been applied in prosthetics due to their similarity to the human musculoskeletal system in terms of interaction and joint stiffness regulation. Recent studies report applications of VSAs in standalone VS prosthetic joints, including ankle-foot~\cite{shepherd2017vspa, glanzer2018design, rogers2023design}, hand~\cite{hocaoglu2022design}, wrist~\cite{milazzo2024modeling}, and elbow~\cite{baggetta2022design, milazzo2024elbow} prostheses. However, multiple VS joints have never been integrated into a single transhumeral prosthesis.

Wrist prostheses are often significantly streamlined to reduce overall complexity. These prosthetic wrists are typically passive and usually feature a limited number of degrees of freedom (DoFs), prioritizing wrist pronation/supination (wPS), occasionally implementing wrist flexion/extension (wFE), and rarely incorporating radial/ulnar deviation (RUD)~\cite{fan2022prosthetic, bajaj2019state, damerla2022design}.

Developing human-like bionic hands has been a longstanding challenge in upper limb prosthetics research. Due to the remarkable complexity and versatility of human hands, artificial hands inevitably require a trade-off between functionality and complexity. Polyarticulated hand prostheses are sophisticated devices that typically include a set of pre-installed standard grips, each designed for a specific task~\cite{mendez2021current, piazza2019century, saikia2016recent, lan2021next}. However, these dexterous devices are often abandoned due to their increased weight, cognitive load, and the frustration of selecting the desired grasping strategy~\cite{smail2021comfort, cordella2016literature}. In contrast, soft hands are easier to use, as their compliant designs allow passive adaptation to object shapes. Additionally, some can modulate stiffness to enhance precision and grasp strength, utilizing technologies such as pneumatic actuators, shape memory alloys, or cable-driven actuation~\cite{li2022soft}.

Bionic limb replacement becomes more challenging as the level of amputation grows, and prosthetic devices must implement a higher number of motor functions while respecting strict morphological (i.e., size and weight) and functional (i.e., range of motion, torque, speed, and battery life) requirements~\cite{milazzo2024elbow,milazzo2024modeling,nasa_dim,tilley2001measure}.
Consequently, transhumeral prostheses often omit some biological DoFs and leave small space for user-centered customizations due to restrictive design constraints~\cite{lenzi2016ric,bennett2016design,abayasiri2017mobio,fite2007gas,johannes2020modular}.

\section{The V-Soft Pro Platform}\label{sec:V-SoftPro}
\begin{figure*}
    \centering
    \includegraphics[width=0.97\linewidth]{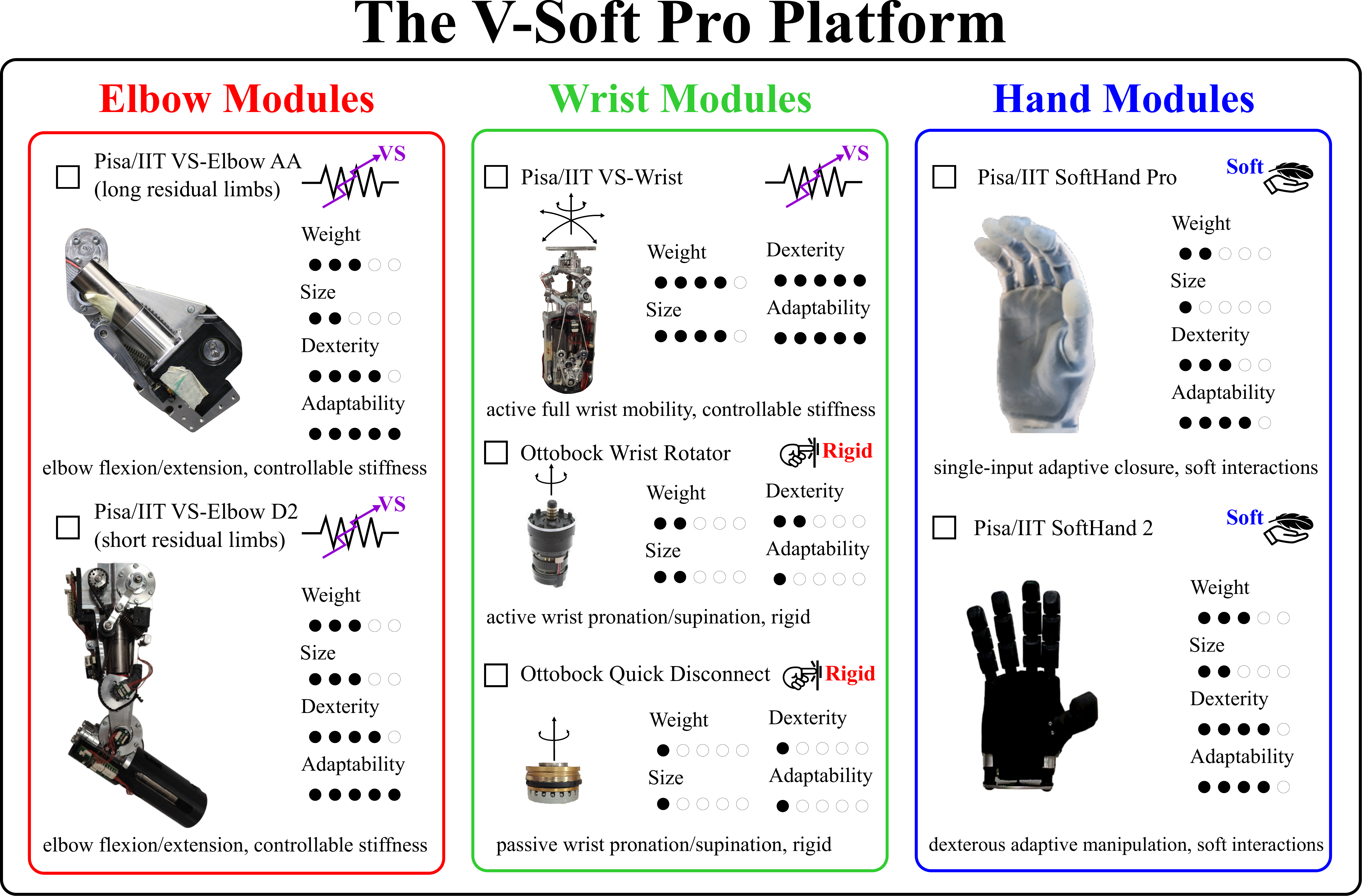}
    \caption{The V-Soft Pro platform. The V-Soft Pro transhumeral prosthesis supports a modular design that can be customized to meet the diverse needs of end-users, including residual limb morphology, required dexterity, comfort with device weight, and available control inputs.}\label{fig:VSoftPro}
\end{figure*}
The V-SP platform features a modular design, outilined in Fig.~\ref{fig:VSoftPro}, allowing users to select the most suitable configuration based on their individual residual limb morphology, number of independent control channels, and daily needs. The platform includes the VS-Elbow, presented in~\cite{milazzo2024elbow}, which offers two different layouts designed to optimize ergonomics and biomimicry for both distal and proximal transhumeral amputations. Additionally, users can choose from three prosthetic wrists, depending on their ability to control numerous DoFs. The V-SP platform includes the VS-Wrist~\cite{milazzo2024modeling}, a prosthetic wrist with 3 active DoFs and controllable stiffness, along with two commercial wrists: the Ottobock Wrist Rotator\footnote{\label{note4}\href{https://www.ottobock.com/en-ex/product/10S17}{Ottobock Wrist Rotator}}
and the Ottobock Quick Disconnect Wrist\footnote{\label{note5}\href{https://shop.ottobock.us/Prosthetics/Upper-Limb-Prosthetics/Myo-Hands-and-Components/Myo-Wrist-Units-and-Rotation/O-B-Disconnect-Adapter/p/11S33\%7E544}{Ottobock Quick Disconnect Wrist}}.
Additionally, the V-SP platform is compatible with two prosthetic hands: the SoftHand Pro and SoftHand 2. This flexibility allows the user to choose from terminal devices with varying levels of dexterity, depending on their myoelectric control skills and daily needs.

\subsection{Elbow Module}\label{sec:Elbow_Module}
The VS-Elbow is a prosthetic elbow featuring redundant actuation and non-linear elastic transmission, which allows for decoupled control of elbow stiffness and position. The VS-Elbow features a single kinematic DoF that actuates elbow flexion/extension (eFE). However, the device incorporates two DC motors, which drive the elbow shaft through an open timing belt. Belt tension is adjusted via a lever arm mechanism that affects the active deflection of extension springs. By leveraging redundant actuation, the VS-Elbow adjusts the active length of the transmission belts, thereby modifying the preload on the elastic transmission and regulating elbow stiffness. This mechanism of stiffness regulation is analogous to the coactivation of agonist-antagonist (AA) muscle groups, such as the biceps and triceps brachii. 
Each DC motor is equipped with a planetary gearbox and a worm gear to amplify output torque, ensuring sufficient performance for standard activities of daily living, such as actively lifting payloads of up to 3 kg. The motor transmission is designed to be non-backdrivable, enabling the VS-Elbow to support payloads and maintain high stiffness passively, which in turn improves battery efficiency. The output joint is decoupled from the motor shaft, ensuring compliant behavior and maintaining the imposed stiffness even when the system is powered off.

The VS-Elbow is available in two different layouts to accommodate the diverse morphologies of residual limbs. The VS-Elbow AA adopts a VSA with an AA configuration, where one elastic actuator represents the triceps brachii and the other represents the biceps brachii. Thus, the net torque generated by one branch relative to the other produces elbow flexion or extension, while their magnitudes determine the joint stiffness. The AA design results in a compact form factor that can be fully contained within the forearm segment of the bionic limb. Due to this feature, the VS-Elbow can be adapted to fit proximal transhumeral amputations while meeting the restrictive functional and morphological requirements of upper limb prostheses, and it can achieve symmetry with the contralateral elbow axis of rotation. However, since the weight of the elbow prosthesis is concentrated in the forearm, the VS-Elbow AA may not offer optimal comfort for individuals with distal transhumeral amputations, as it leaves unused space in the upper limb segment. 

Addressing individuals with short residual limbs, V-Soft Pro features the VS-Elbow D2, a prosthetic elbow that evenly distributes its mass and volume between the upper limb and forearm. This unique design ensures a human-like mass distribution, enhancing ergonomics and promoting natural movement for distal transhumeral amputees~\cite{milazzo2024elbow}. The VS-Elbow D2 is a VSA with an Explicit Stiffness Variation layout, where the two DC motors operate independently to control either the position or the stiffness of the joint. The stiffness motor is located in the upper limb segment and modulates joint stiffness by adjusting the preload of two elastic transmission mechanisms. These non-linear springs are configured to function as an AA system, ensuring symmetry of joint stiffness during both flexion and extension. Conversely, the position motor is located in the forearm segment and drives eFE by rotating the forearm around the fixed elbow shaft. 

\subsection{Wrist Module}\label{sec:Wrist_Module}
The wrist module of V-Soft Pro can be customized to accommodate the final user's available control channels and daily needs. The VS-Wrist is the most advanced wrist module in the platform, featuring 3 active DoFs and user-controllable stiffness. It is the first prosthetic wrist designed to actively restore all human wrist DoFs while providing decoupled stiffness control across a continuous range through modulation of non-linear spring preload. Typically, a 3-DoF VSA requires 6 motors, resulting in significant bulk and weight that would hinder prosthetic applications. However, the VS-Wrist utilizes a hybrid parallel-serial architecture to achieve 3 kinematic DoFs and user-controllable stiffness within a compact design, requiring only 4 motors.
The parallel manipulator (PM) offers a hemispherical range of motion, enabling wrist flexion/extension (wFE) and radial/ulnar deviation (RUD). Although it has two DoFs, the PM is redundantly actuated by three DC motors through a non-linear elastic transmission. This architecture allows modulation of internal forces to the PM, thereby adjusting the preload on the non-linear springs without affecting the wrist equilibrium position, and enabling decoupled stiffness and position control. Similar to the VS-Elbow, each motor transmission incorporates a planetary gearbox and a worm drive, achieving non-backdrivability to maintain static loads without power consumption. Furthermore, the VS-Wrist includes a dedicated motor unit that actuates wrist pronation/supination (wPS) through two opposed universal joints, positioned at the center of the PM.

V-Soft Pro is also compatible with two simpler commercial prosthetic wrists, offering options for users with fewer control inputs or those preferring a lighter device, although at the expense of some functionalities. The Ottobock Wrist Rotator is a commercial myoelectric wrist that supports EMG control of wPS, while wFE and RUD are blocked. The Ottobock Quick Disconnect Wrist is more compact but allows only passive wPS, which the user can actuate using the contralateral arm.

\subsection{Hand Module}\label{sec:Hand_Module}
The SoftHand Pro~\cite{godfrey2018softhand} uses a single motor to actuate 19 DoFs, replicating the kinematics and morphology of a human hand. Each finger consists of three rigid modules, similar to phalanges, with the thumb comprising only two segments. These phalange modules are 3D-printed and available in various sizes. Elastic joints connect consecutive phalanges, ensuring robust and safe interaction with the environment. In this way, rigid links, representing bones, are connected by elastic bands that function like ligaments. A differential transmission evenly distributes contact forces across all fingers, enabling the hand to grasp objects of various shapes.
Thanks to the mechanical intelligence inherent in its design, the SoftHand Pro is intuitive to use, allowing users to execute a variety of grasping strategies that passively adapt to different shapes with a single closing command. For users with advanced EMG control skills, the SoftHand 2 introduces an additional synergistic actuator, expanding the range of grasping, manipulation, and fine gesture capabilities~\cite{della2018toward}. The SoftHand also features a silicone glove or rubber pads, enhancing grasp stability and offering a customizable cosmetic appearance.

\section{Experiments}\label{sec:Experiments}
\begin{figure*}[!t]
    \centering
    \subfloat[]{\includegraphics[width = 0.93\linewidth]{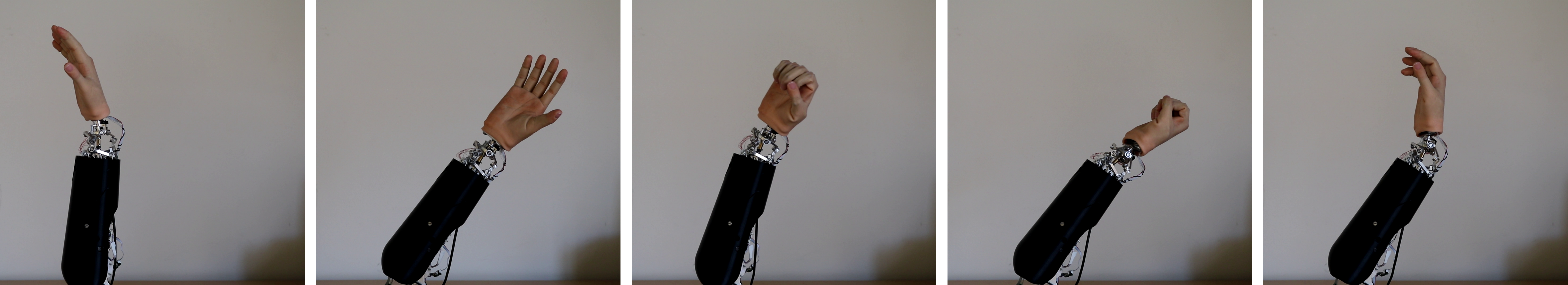}\label{fig:Exp1}}\\
    \subfloat[]{\includegraphics[width = 0.93\linewidth]{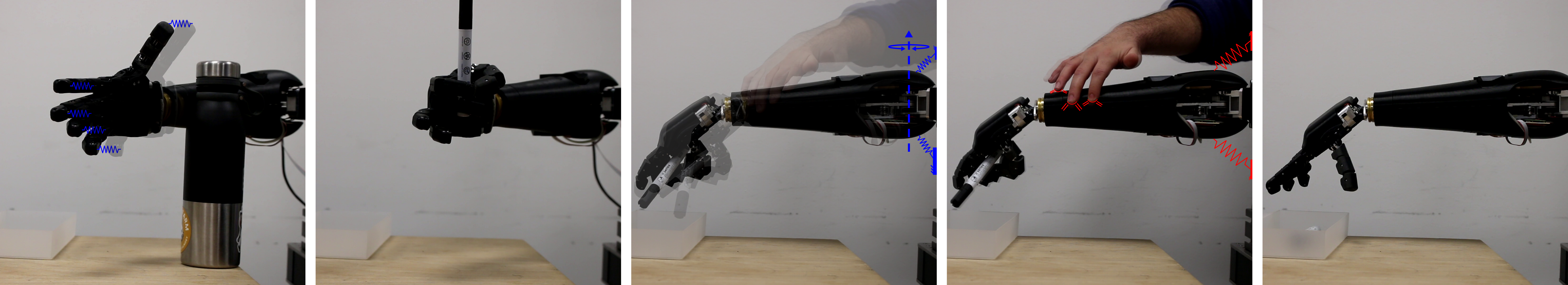}\label{fig:Exp2}}
    \caption{Functional demonstration of two configurations of the V-Soft Pro system. Sequence (a) shows the V-SP platform performing active, simultaneous motion across all its submodules. Sequence (b) demonstrates the benefits of variable stiffness actuation within the transhumeral prosthesis. In the first scene, the prosthesis uses low joint stiffness to navigate an unstructured environment, allowing gentle contact with nearby objects without knocking them over. It then performs a tripod grasp (second scene) followed by wrist pronation. With low elbow stiffness, the prosthesis adapts to external forces without resisting (third scene). However, the V-SP system can increase its stiffness to counter external disturbances (fourth scene) and perform precision tasks (fifth scene).}\label{fig:Experiments}
\end{figure*}
We present the integration of the V-SP platform in two distinct configurations. The first configuration includes the VS-Elbow D2, the VS-Wrist, and the SoftHand Pro, providing a total of 5 DoFs for kinematics and 2 DoFs for stiffness control. The second configuration utilizes the VS-Elbow AA, the Ottobock Wrist Rotator, and the SoftHand 2, offering 4 kinematic DoFs and 1 DoF to control elbow stiffness.

Each motor position is regulated by a dedicated PID controller operating at 200 Hz, which modulates the duty cycle of a 20 kHz PWM signal. The system is powered by a single 24 V commercial battery. An external laptop, running a Matlab Simulink graphic user interface, enables the user to set position and stiffness references for each prosthetic joint, as well as to monitor encoder and EMG signals in real-time.

We report preliminary lab tests demonstrating the varying degrees of prosthesis functionality achievable through customization of the modular V-SP platform. A comprehensive technical characterization of the V-SP platform submodules can be found in~\cite{milazzo2024elbow,milazzo2024modeling,godfrey2018softhand,della2018toward}.
Fig.~\ref{fig:Experiments} presents photo sequences extracted from the multimedia material accompanying this paper. Fig.~\ref{fig:Exp1} illustrates the first configuration of the V-SP system during coordinated active motion of all prosthetic joints, highlighting its high wrist mobility.
Fig.~\ref{fig:Exp2} depicts the second configuration of the V-SP system, where the prosthesis adjusts its elbow stiffness to adapt to different tasks. This adaptability suggests the enhanced versatility of V-SP prostheses in unstructured environments. The V-SP transhumeral prosthesis can reduce elbow stiffness to accommodate obstacles along its path and comply with external forces, enabling gentle interactions. Conversely, it increases elbow stiffness to reject disturbances and support precision tasks.

\section{Discussion}\label{sec:Discussion}
Current robotic prostheses significantly enhance the quality of life for upper limb amputees. However, they typically incorporate simplified motor functions and lack the natural compliance of human limbs. Replicating the controllable stiffness of the human musculoskeletal system in a prosthetic device would enable more human-like interactions, thus increasing acceptance and social integration. Moreover, compliant bionic limbs improve safety by absorbing unintended impacts that can arise from the absence of proprioceptive feedback or from errors in the operation of the prosthetic limb.
Furthermore, VS prostheses can adapt their behavior to various tasks encountered in daily life. As shown in Fig.~\ref{fig:Exp2}, soft joints enable the exploitation of environmental constraints, ensuring safe interactions in cluttered environments by adapting to obstacles and preventing damage. Meanwhile, high elbow stiffness is crucial for rejecting disturbances during precision tasks. While this preliminary experiment suggests that VS prosthetic limbs may improve users' quality of life, further testing with prosthesis users is necessary to evaluate their real-world functionality.

The V-SP platform offers a customizable transhumeral prosthesis that meets diverse user specifications, considering both ergonomics and the number of controllable motions. Future work will focus on tailoring the EMG control of the V-SP system to the individual user. One possibility is to adopt existing EMG-based VS controllers from the literature, which regulate the stiffness and position of single-DoF joints~\cite{sensinger2008user, capsi2020exploring, ferrante2024toward}, and extend these approaches to sequential control, commonly used in multi-DoF prostheses. In this scenario, the user utilizes a switching signal to toggle control between different prosthetic joints, such as rapid coactivations, signals from auxiliary EMG recording sites, or even a physical switch activated by the contralateral arm. While sequential control is easy to implement, it may result in unnatural motion and unintuitive control as the number of DoFs increases, limiting the usability of advanced prostheses. Key enablers of this technology may include surgical procedures that increase the number of independent biosignals (e.g., Targeted Muscle Reinnervation and Regenerative Peripheral Nerve Interfaces), cutting-edge EMG recording technologies (such as high-density EMG and implantable electrodes), and advanced neural decoding algorithms~\cite{mendez2021current, huang2024integrating, farina2023toward, jyothish2024survey}.

\section{Conclusion}\label{sec:Conclusion}
Current robotic prostheses lack the innate compliance and variable stiffness of human limbs, leading to unnatural interactions with people and objects, while compromising versatility and social engagement.
To address these issues, this paper presents V-Soft Pro, a modular platform for a transhumeral prosthesis with user-controllable stiffness. V-Soft Pro features a user-centered design, enabling customization of its components based on the user's level of amputation, daily needs, and myoelectric control skills. Recognizing the diversity among users, the modular design offers various configurations to optimize comfort and usability for each individual. The prosthetic joints integrate VSAs to replicate the adaptive behavior of human limbs, providing human-like dynamics and continuous joint stiffness regulation.
The V-Soft Pro system has the potential to significantly enhance the user experience by improving the natural feel of bionic limbs and offering customized prosthetic solutions to individual needs. However, further research is necessary to accurately decode user intent, enabling more natural operation of the prosthesis and maximizing its capabilities. Clinical trials with prosthesis users are needed to assess the functionality of user-controllable stiffness in prosthetic devices during ADLs and social interactions.

\section*{Acknowledgment}
The authors would like to thank Manuel Barbarossa, Vinicio Tincani, Mattia Poggiani, Marina Gnocco, Cristiano Petrocelli, Emanuele Sessa, and Octavie Somoza for their fundamental work on the hardware implementation of the prototypes.

\bibliography{References}
\bibliographystyle{IEEEtran}

\end{document}